\documentclass[sigconf,natbib=false]{acmart}
\AtBeginDocument{%
  }

\setcopyright{acmlicensed}
\copyrightyear{2026}
\acmYear{2026}
\setcopyright{cc}
\setcctype{by}
\acmConference[SAC '26]{The 41st ACM/SIGAPP Symposium on Applied Computing}{March 23--27, 2026}{Thessaloniki, Greece}
\acmBooktitle{The 41st ACM/SIGAPP Symposium on Applied Computing (SAC '26), March 23--27, 2026, Thessaloniki, Greece}
\acmPrice{}
\acmDOI{10.1145/3748522.3780009}
\acmISBN{979-8-4007-2294-3/2026/03}

\RequirePackage[
  datamodel=acmdatamodel,
  style=acmnumeric,
  ]{biblatex}

\usepackage{graphicx}
\usepackage{subcaption}

\begin{document}

\title{Explainable Anomaly Detection for Industrial IoT Data Streams}
% Please make sure that the short title does not exceed the width of one column
\renewcommand{\shorttitle}{Explainable Anomaly Detection for Industrial IoT Data Streams}


\author{Ana Rita Paupério, Diogo Risca, Afonso Lourenço, Goreti Marreiros}
\affiliation{%
  \institution{GECAD/LASI, Polytechnic of Porto}
  \city{Porto, 4249-015}
  \country{Portugal}}
\email{{armpo,difri,fonso,mgt}@isep.ipp.pt}

\author{Ricardo Martins}
\affiliation{%
  \institution{SISTRADE Software Consulting, S.A.}
  \city{Porto, 4250-380}
  \country{Portugal}}
\email{ricardo.martins@sistrade.com}

%This command displays author info in page headers
% Please use the following convention:
% One author: J. Smith
% Two authors: J. Smith and I. Jones
% Three and more authors: J. Smith et al.
\renewcommand{\shortauthors}{Paupério et al.}

\begin{abstract}
Industrial maintenance is being transformed by the Internet of Things and edge computing, generating continuous data streams that demand real-time, adaptive decision-making under limited computational resources. While data stream mining (DSM) addresses this challenge, most methods assume fully supervised settings, yet in practice, ground-truth labels are often delayed or unavailable. This paper presents a collaborative DSM framework that integrates unsupervised anomaly detection with interactive, human-in-the-loop learning to support maintenance decisions. We employ an online Isolation Forest and enhance interpretability using incremental Partial Dependence Plots and a feature importance score, derived from deviations of Individual Conditional Expectation curves from a fading average, enabling users to dynamically reassess feature relevance and adjust anomaly thresholds. We describe the real-time implementation and provide initial results for fault detection in a Jacquard loom unit. Ongoing work targets continuous monitoring to predict and explain imminent bearing failures.
\end{abstract}

\begin{CCSXML}
<ccs2012>
   <concept>
       <concept_id>10010147.10010257.10010282.10010284</concept_id>
       <concept_desc>Computing methodologies~Online learning settings</concept_desc>
       <concept_significance>500</concept_significance>
   </concept>
   <concept>
       <concept_id>10010147.10010257.10010258.10010262.10010279</concept_id>
       <concept_desc>Computing methodologies~Learning under covariate shift</concept_desc>
       <concept_significance>500</concept_significance>
   </concept>
   <concept>
       <concept_id>10010147.10010257.10010293.10010319</concept_id>
       <concept_desc>Computing methodologies~Learning latent representations</concept_desc>
       <concept_significance>500</concept_significance>
   </concept>
</ccs2012>
\end{CCSXML}

\ccsdesc[500]{Computing methodologies~Online learning settings}
\ccsdesc[500]{Computing methodologies~Learning under covariate shift}
\ccsdesc[500]{Computing methodologies~Learning latent representations}

\iffalse

\begin{CCSXML}
<ccs2012>
 <concept>
  <concept_id>00000000.0000000.0000000</concept_id>
  <concept_desc>Do Not Use This Code, Generate the Correct Terms for Your Paper</concept_desc>
  <concept_significance>500</concept_significance>
 </concept>
 <concept>
  <concept_id>00000000.00000000.00000000</concept_id>
  <concept_desc>Do Not Use This Code, Generate the Correct Terms for Your Paper</concept_desc>
  <concept_significance>300</concept_significance>
 </concept>
 <concept>
  <concept_id>00000000.00000000.00000000</concept_id>
  <concept_desc>Do Not Use This Code, Generate the Correct Terms for Your Paper</concept_desc>
  <concept_significance>100</concept_significance>
 </concept>
 <concept>
  <concept_id>00000000.00000000.00000000</concept_id>
  <concept_desc>Do Not Use This Code, Generate the Correct Terms for Your Paper</concept_desc>
  <concept_significance>100</concept_significance>
 </concept>
</ccs2012>
\end{CCSXML}

\ccsdesc[500]{Do Not Use This Code~Generate the Correct Terms for Your Paper}
\ccsdesc[300]{Do Not Use This Code~Generate the Correct Terms for Your Paper}
\ccsdesc{Do Not Use This Code~Generate the Correct Terms for Your Paper}
\ccsdesc[100]{Do Not Use This Code~Generate the Correct Terms for Your Paper}

\fi

\keywords{Explainable AI (XAI), Edge Computing, Online Anomaly Detection}

\maketitle

\begin{figure*}[ht]
    \centering
    \begin{subfigure}{0.46\textwidth}
        \centering
        \includegraphics[width=\textwidth]{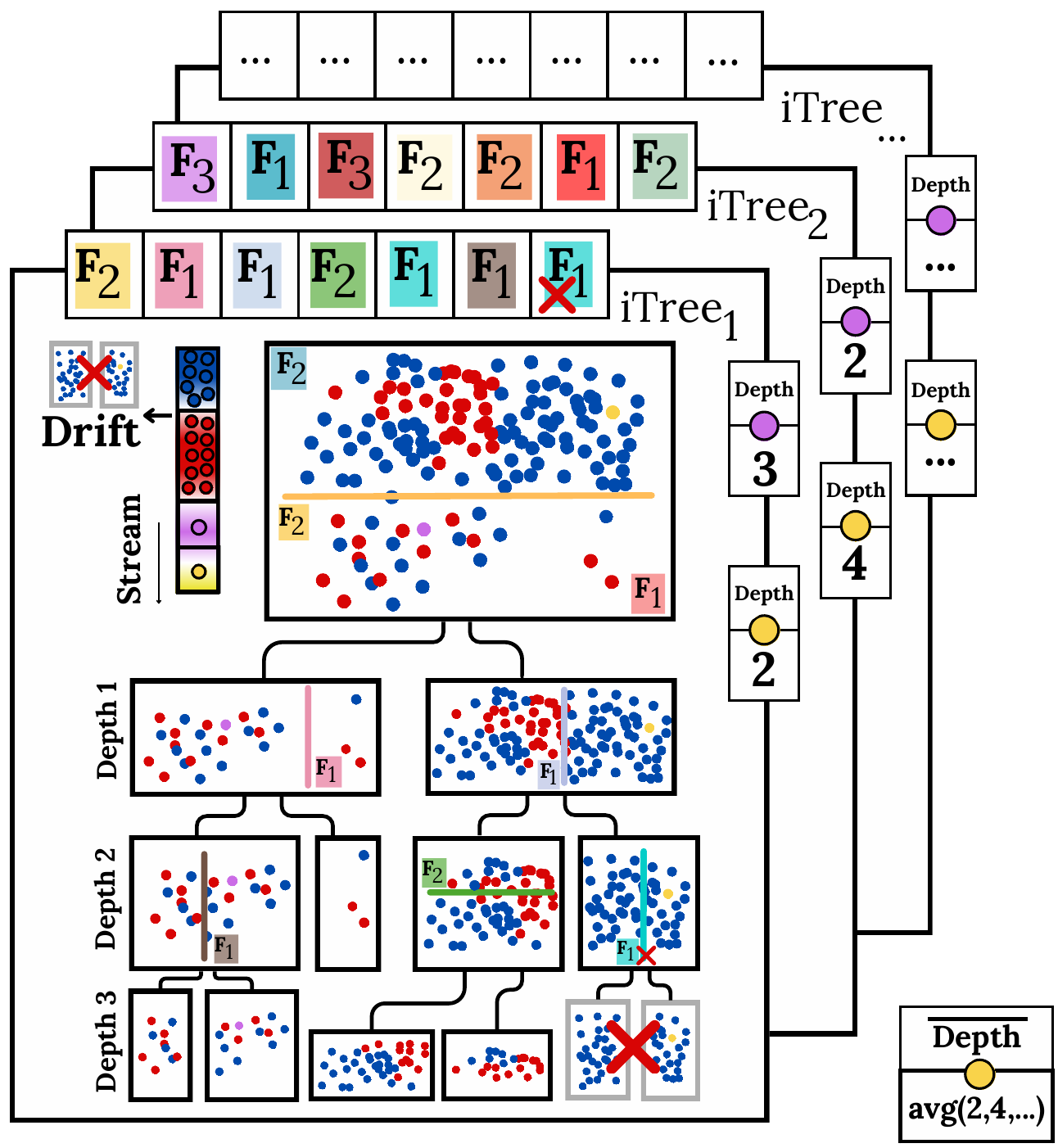}
        \caption{Each Onl-iTree performs random splits that are pruned over time based on region bin counts. In iTree-1, for example, outdated points (blue) decrease counts and trigger a merge under a recent split (turquoise). The anomaly score is computed from the average path lengths given the trees structure at the instance’s arrival time.}
        \label{fig:iforest1}
    \end{subfigure}
    \hfill
    \begin{subfigure}{0.42\textwidth}
        \centering
        \includegraphics[width=\textwidth]{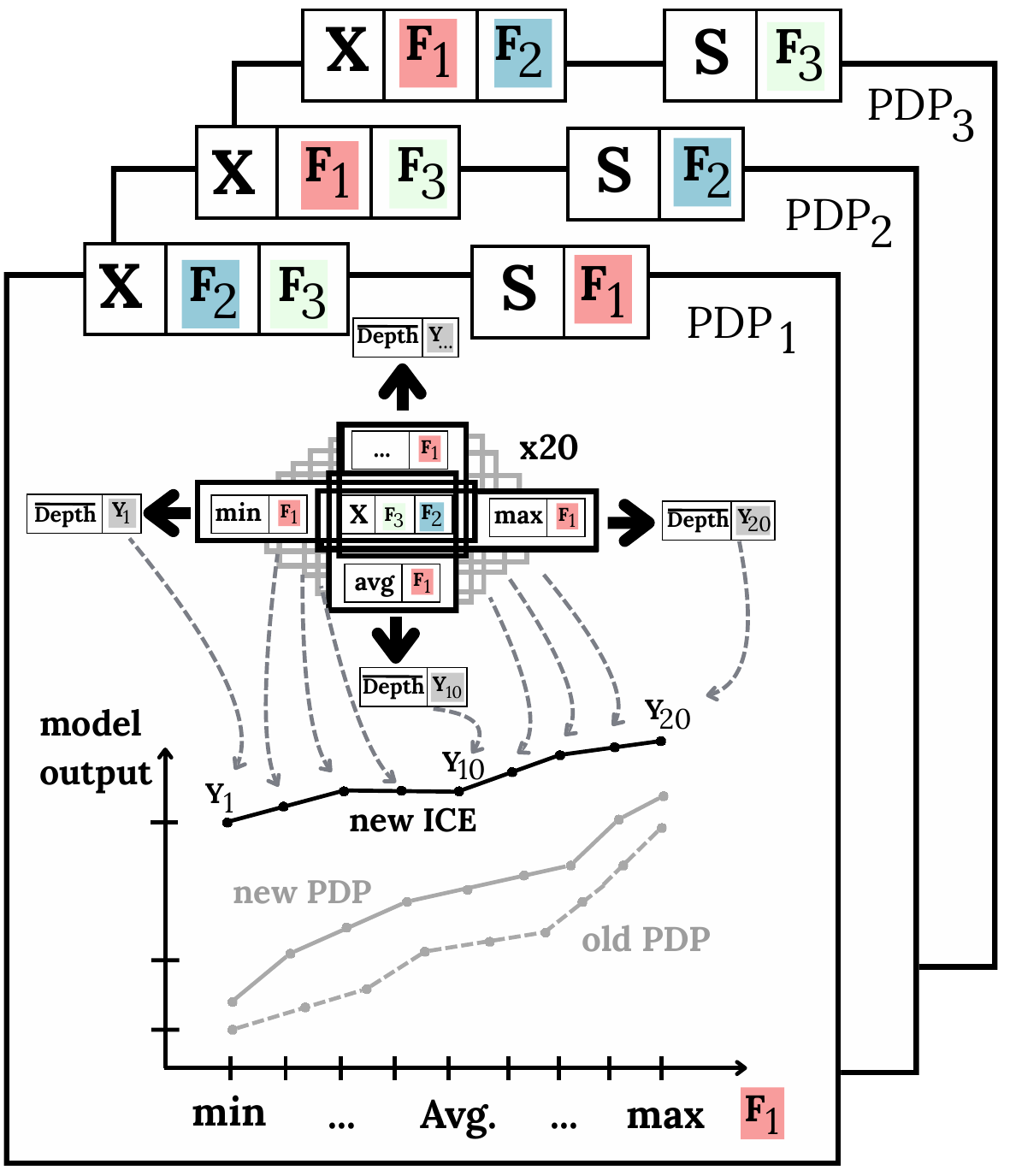}
        \caption{For each feature, a new ICE curve is computed by evaluating the Onl-iForest depth over 20 evenly spaced points between the current minimum and maximum values, keeping all other features fixed at the incoming instance. The iPDP is the exponential moving average of these ICE curves.}
        \label{fig:iforest2}
    \end{subfigure}
    \caption{Online iForest with adaptive pruning (left) and incremental marginal feature effects via streaming ICE curves (right).}
\end{figure*}

\section{Introduction}

Industrial maintenance has evolved rapidly as IoT, edge computing, and autonomous systems produce continuous, high-speed data streams \cite{lourencco2025dfdt}. These require machine learning models to operate in real time, adapt to concept drifts, and handle resource constraints \cite{risca2025continual}. To address this, data stream mining (DSM) methods have been proposed, though most rely on fully supervised adaptation  \cite{neves2025online,lourencco2025device}.

However, in real-world scenarios, it is unrealistic to provide ground-truth labels immediately upon prediction. We argue that in most high-latency and high-drifting scenarios, it might be better to assume that no labels will ever be available. In this case, framing a classification problem as \textbf{novelty detection (ND)}, e.g. with MINAS \cite{faria2016minas}, can be advantageous if one can frame events of interest as a cohesive agglomeration of anomalies over time. Alternatively, if one can frame events of interest as \textit{few} and \textit{different}, framing a classification problem as \textbf{anomaly detection (AD)}, e.g. with an Online Isolation Forest (Onl-iForest) \cite{pmlr-v235-leveni24a}, can be advantageous. However, since these unsupervised methods rely solely on statistical properties, both detected novelties and anomalies may not always correspond to semantically relevant events for the operator using the system. To address this, one needs to guide the user to intentionally act on the DSM system, providing richer information for learning. In this regard, selecting an appropriate strategy is necessarily dependent on whether the DSM system automates or collaborates \cite{natarajan2025human}:

\begin{itemize}
    \item \textbf{DSM automation requires machine teaching (MT)}.  In a manufacturing plant, a DSM system can predict the remaining useful life (RUL) of a component to automatically schedule maintenance and order materials, while the labeling naturally arises from the actual lifespan. When RUL spans hours or days, supervised DSM is feasible. For longer periods, delayed labels make full supervision impractical. In such cases, it is key to engage humans not only as oracles to label uncertain samples, but also as teachers, providing advice such as feature selection, inductive biases, or constraints on model structure to guide learning. For example, an engineer might be prompted to highlight subtle degradation patterns in sensor data that the model would otherwise misinterpret, accelerating learning and reducing prediction errors in RUL.
    \item \textbf{DSM collaboration requires interactive learning (IL)}. Here, the DSM system recommends actions that are accepted based on human judgment. When an engineer inspects and confirms wear or normal operation, the result becomes the label. Labeling delay thus depends on inspection frequency and fault occurrence. In such cases, as the user is in control and unlikely to trust misaligned recommendations, the model must continuously justify its suggestions, presenting uncertainty estimates (“I know what I don’t know”) and explaining its reasoning (“I know why I decided this”).
\end{itemize}

With these notions in mind, we propose a \textbf{collaborative DSM system} that recommends whether a Jacquard loom machine requires inspection, via an \textbf{AD} method, highlighting \textbf{IL} components. Specifically, we employ incremental Partial Dependence Plots (iPDPs) to examine how the Onl-iForest’s anomaly scores change when observations are forced to take fixed values for individual features.

\section{Anomaly detection}

For AD, we adopt Onl-iForest \cite{pmlr-v235-leveni24a}. Unlike the batch-mode iForest, Onl-iForest maintains tree structures incrementally through a sliding window, adapting to concept drift in streaming environments. This allows for fast and efficient anomaly detection with constrained memory while preserving modeling fidelity.

Onl-iForest maintains a forest $\mathcal{F} = \{T_1, \dots, T_\tau \}$ of trees updated in real time through a sliding window $W$ of the $\omega$ most recent points. Each tree partitions $\mathbb{R}^d$ into axis-aligned bins. The nodes are defined by their count $h$ and region $\mathcal{R}$. If $h$ exceeds a threshold (increasing with depth), the node splits along a randomly chosen feature and value. As new samples arrive, bin counts are updated; with outdated points decrementing counts, potentially triggering merges of underpopulated subtrees. Figure~\ref{fig:iforest1} illustrates this: the child nodes of the turquoise split are pruned and merged into a parent node. To compute an anomaly score for a new instance, each tree reports the depth of the leaf that contains the point. The final score is calculated using Equation \ref{eq:1}, where $E(\mathcal{D})$ is the average depth and $\eta$ is a constant that regulates the normalization of the tree depth, typically set relative to the size of the sliding window, to maintain consistent scores \cite{pmlr-v235-leveni24a}. 
\begin{equation} \label{eq:1}
    s_t = 2^{-E(\mathcal{D}) / \log_2(\omega / \eta)}
\end{equation}
This scoring mechanism relies on the principle that anomalies are isolated more quickly and appear in shallower nodes.

\section{Interactive learning}

IL enables greater human involvement and shared control in the learning process \cite{natarajan2025human}. In AD, it is essential to continuously redefine what constitutes an “anomalous’’ threshold, to update assumptions about the “normal’’ data-generating process, and to reassess which features should drive predictions. However, many state-of-the-art AD methods, such as Onl-iForest, are too complex to be interpreted in a way that allows humans to effectively understand and influence the behavior of the model. To address this limitation, we employ iPDPs, which examine the response of the model when all observations are forced to take a fixed value for a given feature \cite{muschalik2023ipdp}. 

To handle the streaming setting, iPDP updates feature effect estimates as new data arrives by combining the current model evaluations with previous estimates using an exponential moving average of Individual Conditional Expectation (ICE) curves. Figure~\ref{fig:iforest2} illustrates this: we evaluate the ICE of each feature at 20 points, forming a smooth adaptive grid whose locations follow the evolving feature distribution. When the range of the feature changes significantly, the method dynamically computes the minimum and maximum using a sampling procedure, ensuring that the evaluation points remain evenly distributed.

Beyond visual analysis of the marginal feature effect in PDPs (e.g. whether it is monotonic, or non-linear), the ICE curves can also be aggregated into a Feature Importance (FI) score by quantifying how individual effects deviate from the corresponding average curve.

\begin{figure}[h]
    \centering
    \includegraphics[width=0.44\textwidth]{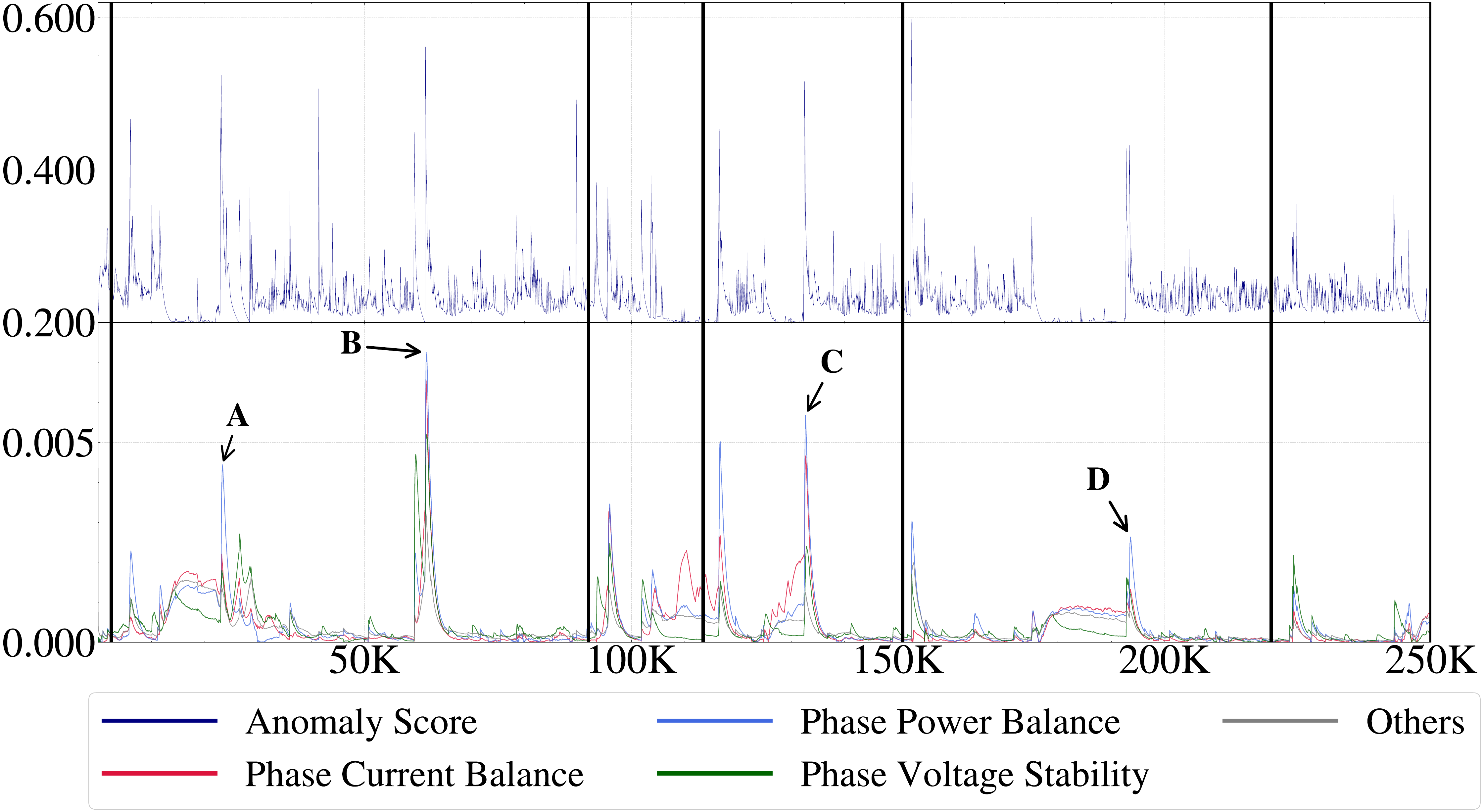}
    \caption{Anomaly and FI scores. Horizontal marks indicate production runs. Points \textit{A}, \textit{B}, \textit{C}, \textit{D} indicate events of interest.}
    \label{fig:anomaly_scores_a}
\end{figure}

\section{Discussion}

This case study presents preliminary results for predictive maintenance of a Jacquard loom, a weaving machine that produces complex patterns by individually controlling warp threads. While the loom detects electrical faults, mechanical issues, such as bearing degradation, often remain unnoticed until they become critical. To identify emerging faults, the DSM system monitors a set of features: \textit{Phase Current Balance}, \textit{Phase Voltage Stability}, \textit{Phase Power Balance}, \textit{Thermal Stress}, \textit{Current THD Spread}, \textit{Voltage Quality}, \textit{Phase Reactive Flow}, \textit{Phase Efficiency Ratio}, \textit{Phase Apparent Power}.

\begin{figure}[h]
    \centering
    \includegraphics[width=0.47\textwidth]{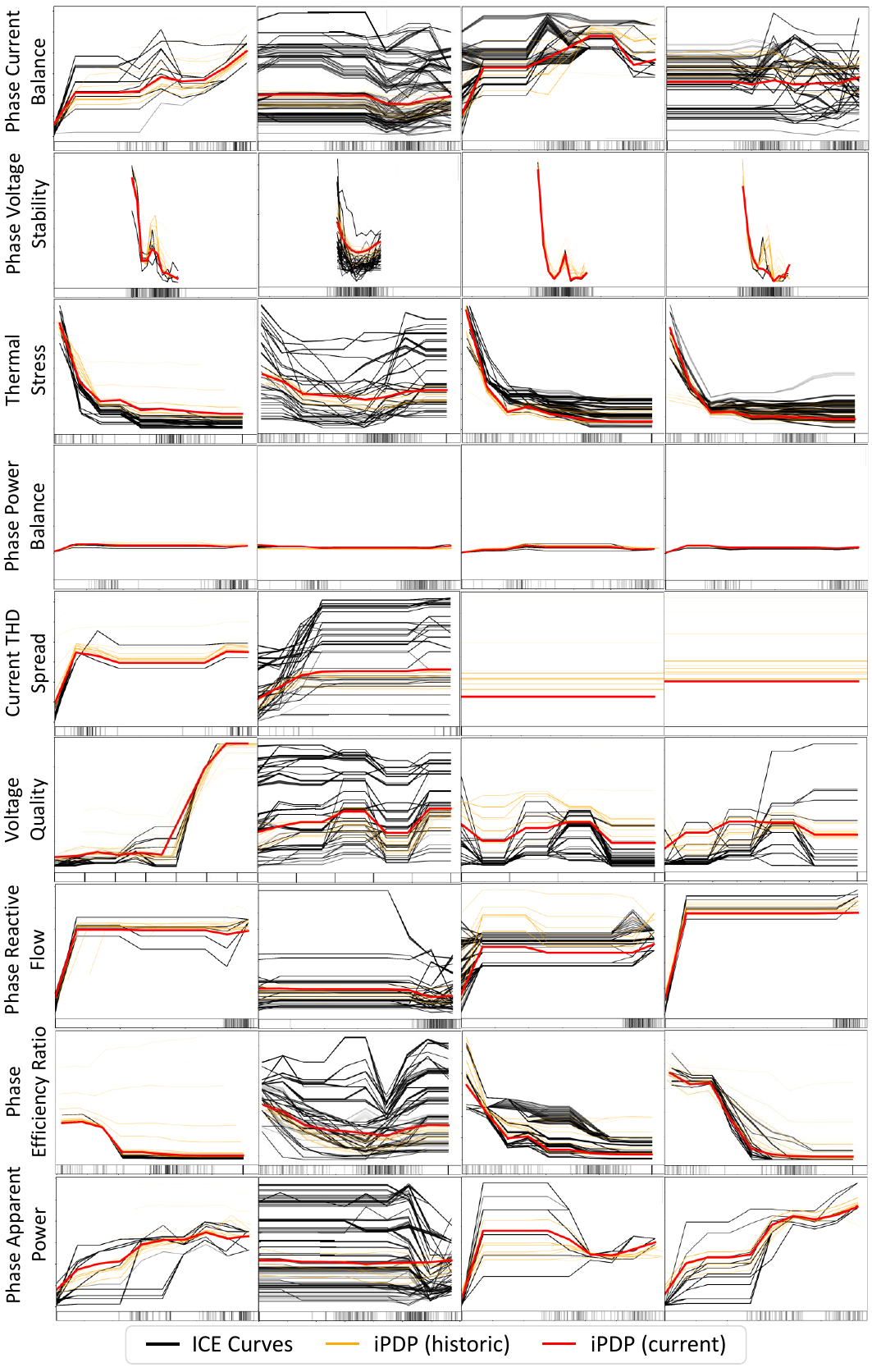}
    \caption{Partial dependence plots for the nine features (rows) across four anomalies \textit{A}, \textit{B}, \textit{C}, \textit{D} (columns).}
    \label{fig:pdp_anomalies}
\end{figure}

Figure \ref{fig:anomaly_scores_a} shows how the FI and anomaly scores evolve over the monitoring period. As anticipated, inspecting the anomaly scores alone provides little actionable insight. In contrast, the FI of the top three variables (\textit{Phase Current Balance}, \textit{Phase Voltage Stability}, and \textit{Phase Power Balance}) correlate well not only with the abrupt shifts caused by changes in raw materials across between production runs (horizontal marks), but also with four additional spikes (\textit{A}, \textit{B}, \textit{C}, \textit{D}). Interpreting these requires examining not only the magnitude but also the patterns of feature contributions. Figure \ref{fig:pdp_anomalies} presents snapshot PDPs and fading ICE curves at these anomaly times. Anomalies \textit{A} and \textit{C} show rising \textit{Phase Current Balance Index} and \textit{Phase Apparent Power Index}, alongside sudden drops in \textit{Phase Voltage Stability Index} and \textit{Phase Power Balance Index}, suggesting motor stress and voltage imbalance. Both also display sharp increases in \textit{Phase Reactive Flow Index}, indicating transient reactive power fluctuations. \textit{A} further exhibits increases in \textit{Current THD Spread} and \textit{Voltage Quality Score}, pointing to harmonic disturbances and declining power quality. In contrast, \textit{B} remains largely flat across features, reflecting a stable but potentially latent state with low mechanical stress. \textit{D} shows mixed behavior, with sudden decreases in \textit{Phase Voltage Stability Index}, \textit{Phase Power Balance Index}, and \textit{Phase Efficiency Ratio}, coupled with spikes in \textit{Phase Reactive Flow Index} and \textit{Phase Apparent Power Index}, signaling abrupt load changes and efficiency loss.

It is important to note that data sparsity in certain feature regions can bias these PDP interpretations. Ongoing work focuses on imminent bearing failures, anticipating costly downtime.

\section*{Acknowledgments}

Funded by the Portuguese Republic’s Recovery and Resilience Plan, under PRODUTECH R3, and FCT, under project UIDP/00760/2020, and EIT Manufacturing Doctoral School, KAVA 23362.

\printbibliography

@String{Springer = "Springer-Verlag" }

@inproceedings{muschalik2023ipdp,
author="Muschalik, Maximilian
and Fumagalli, Fabian
and Jagtani, Rohit
and Hammer, Barbara
and H{\"u}llermeier, Eyke",
title="iPDP: On Partial Dependence Plots in Dynamic Modeling Scenarios",
booktitle="Explainable Artificial Intelligence",
year="2023",
publisher="Springer",
pages="177--194"
}

@article{neves2025online,
  title={Online hierarchical partitioning of the output space in extreme multi-label data stream},
  author={Neves, Lara and Louren{\c{c}}o, Afonso and Cano, Alberto and Marreiros, Goreti},
  journal={arXiv e-prints},
  pages={arXiv--2507},
  year={2025}
}

@inproceedings{pmlr-v235-leveni24a,
  title={Online isolation forest},
  author={Leveni, Filippo and Cassales, Guilherme Weigert and Pfahringer, Bernhard and Bifet, Albert and Boracchi, Giacomo},
  booktitle={Proceedings of the 41st International Conference on Machine Learning},
  pages={27288--27298},
  year={2024}
}

@article{lourencco2025device,
  title={On-device edge learning for IoT data streams: a survey},
  author={Louren{\c{c}}o, Afonso and Rodrigo, Jo{\~a}o and Gama, Jo{\~a}o and Marreiros, Goreti},
  journal={arXiv preprint arXiv:2502.17788},
  year={2025}
}

@article{risca2025continual,
  title={Continual learning for rotating machinery fault diagnosis with cross-domain environmental and operational variations},
  author={Risca, Diogo and Louren{\c{c}}o, Afonso and Marreiros, Goreti},
  journal={arXiv preprint arXiv:2504.10151},
  year={2025}
}

@inproceedings{natarajan2025human,
  title={Human-in-the-loop or AI-in-the-loop? Automate or Collaborate?},
  author={Natarajan, Sriraam and Mathur, Saurabh and Sidheekh, Sahil and Stammer, Wolfgang and Kersting, Kristian},
  booktitle={Proceedings of the AAAI Conference},
  volume={39},
  pages={28594--28600},
  year={2025}
}

@article{faria2016minas,
  title={MINAS: multiclass learning algorithm for novelty detection in data streams},
  author={Faria, Elaine Ribeiro de and Gama, Jo{\~a}o and Carvalho, Andr{\'e} Carlos Ponce de Leon Ferreira},
  journal={Data Mining and Knowledge Discovery},
  volume={30},
  number={3},
  pages={640--680},
  year={2016},
  publisher={Springer}
}

@article{lourencco2025dfdt,
  title={Dfdt: Dynamic fast decision tree for iot data stream mining on edge devices},
  author={Louren{\c{c}}o, Afonso and Rodrigo, Jo{\~a}o and Gama, Jo{\~a}o and Marreiros, Goreti},
  journal={arXiv preprint arXiv:2502.14011},
  year={2025}
}

%\appendix

\end{document}